\documentclass{article} 
\usepackage{iclr2024_conference,times}

\usepackage{amsmath,amsfonts,bm}

\def\eqref#1{equation~\ref{#1}}

\def\1{\bm{1}}

\DeclareMathAlphabet{\mathsfit}{\encodingdefault}{\sfdefault}{m}{sl}
\SetMathAlphabet{\mathsfit}{bold}{\encodingdefault}{\sfdefault}{bx}{n}

\usepackage{hyperref}
\usepackage{url}

\usepackage{amsmath,amsfonts}
\usepackage{amssymb}
\usepackage{algorithmic}
\usepackage{algorithm}
\usepackage{array}
\usepackage{graphicx}
\usepackage{xspace}
\usepackage{ragged2e}
\usepackage{blindtext}
\usepackage{multirow}

\usepackage{booktabs}
\usepackage{float}
\usepackage{xcolor}
\usepackage{cuted}
\usepackage{colortbl}
\usepackage{caption}

\graphicspath{ {./Figure/} }

\title{CEDNet: A Cascade Encoder-Decoder Network for Dense Prediction}

\author{Gang Zhang$^{1}$, Ziyi Li$^{4}$, Chufeng Tang$^{1}$, Jianmin Li$^{1}$, Xiaolin Hu$^{1,2,3}$\thanks{Corresponding Author}\\
$^{1}$Department of Computer Science and Technology, Institute for AI, \\ BNRist, THU-Bosch JCML Center, Tsinghua University, Beijing 100084, China \\
$^{2}$Tsinghua Laboratory of Brain and Intelligence (THBI), \\ IDG/McGovern Institute for Brain Research, Tsinghua University, Beijing 100084, China \\
$^{3}$Chinese Institute for Brain Research (CIBR), Beijing 100010, China \\
$^{4}$Huazhong University of Science and Technology, Wuhan 430074, China \\
{\texttt\small \{zhang-g19,tcf18\}@mails.tsinghua.edu.cn, liziyi@hust.edu.cn,}\\
{\texttt\small \{lijianmin,xlhu\}@mail.tsinghua.edu.cn} \\
Code: {\color{red}\href{https://github.com/zhanggang001/CEDNet}{https://github.com/zhanggang001/CEDNet}}
}

\newcommand{\eg}{{\emph{e.g.}}, }
\newcommand{\ie}{{\emph{i.e.}}, }

\iclrfinalcopy 
\begin{document}

\maketitle

\begin{abstract}
  Multi-scale features are essential for dense prediction tasks, such as object detection, instance segmentation, and semantic segmentation. The prevailing methods usually utilize a classification backbone to extract multi-scale features and then fuse these features using a lightweight module (\eg the fusion module in FPN and BiFPN, two typical object detection methods). However, as these methods allocate most computational resources to the classification backbone, the multi-scale feature fusion in these methods is delayed, which may lead to inadequate feature fusion. While some methods perform feature fusion from early stages, they either fail to fully leverage high-level features to guide low-level feature learning or have complex structures, resulting in sub-optimal performance. We propose a streamlined cascade encoder-decoder network, dubbed CEDNet, tailored for dense \mbox{prediction} tasks. All stages in CEDNet share the same encoder-decoder structure and perform multi-scale feature fusion within the decoder. A hallmark of CEDNet is its ability to incorporate high-level features from early stages to guide low-level feature learning in subsequent stages, thereby enhancing the effectiveness of multi-scale feature fusion. We explored three well-known encoder-decoder structures: Hourglass, UNet, and FPN. When integrated into CEDNet, they performed much better than traditional methods that use a pre-designed classification backbone combined with a lightweight fusion module. Extensive experiments on object detection, instance segmentation, and semantic segmentation demonstrated the effectiveness of our method. The code has been released.
\end{abstract}

\section{Introduction}\label{introduction}

In recent years, both convolutional neural networks (CNNs) and transformer-based networks have achieved remarkable results in various computer vision tasks, including image classification, object detection, and semantic segmentation. In image classification, the widely-used CNNs~\citep{AlexNet,VGG,GoogleNet,ResNet,convnext} as well as the recently developed transformer-based networks~\citep{swin,FocalTransformer,CSWin,HiVIT} generally follow a sequential architectural design. They progressively reduce the spatial size of feature maps and make predictions based on the coarsest scale of features. However, in dense prediction tasks,  such as object detection and instance segmentation, the need for multi-scale features arises to accommodate objects of diverse sizes. Therefore, effectively extracting and fusing multi-scale features becomes essential for the success of these tasks~\citep{MaskRCNN,RetinaNet,FCOS,UPerNet,RefineMask,BPR}.

\begin{figure*}[ht]
  \centering
  \includegraphics[width=1.0\textwidth]{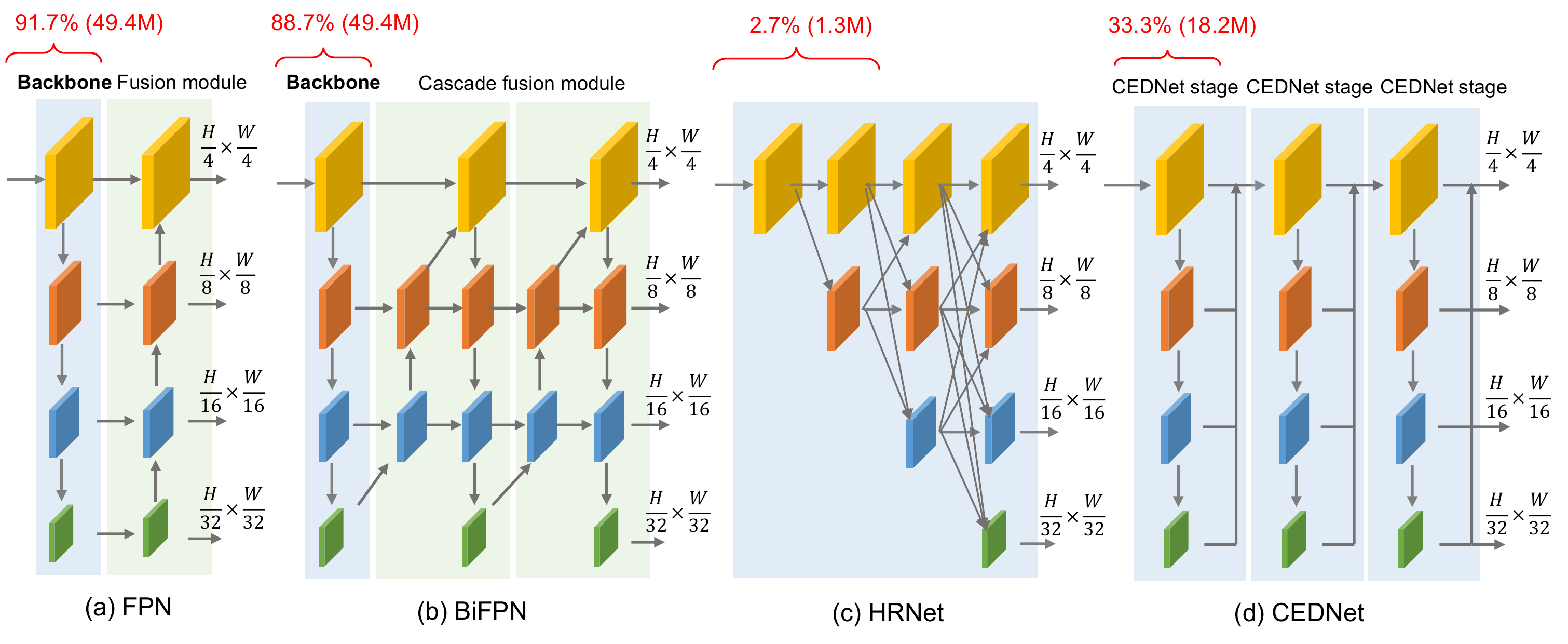}
  \vspace*{-7mm}
  \caption{Comparison among FPN, BiFPN, HRNet, and our CEDNet. {\small $H\times W$} denotes the spatial size of the input image. At the top of each panel, the percentage indicates the time to perform the first multi-scale feature fusion, while the number in bracket is the number of parameters of the selected part. For these calculations, we take ConvNeXt-S as the backbone of FPN and BiFPN. While we illustrate a CEDNet of four scales here for a clearer comparison, it's noteworthy that in our actual implementation, feature maps with a resolution of $\frac{H}{4}\times\frac{W}{4}$ are not included in the CEDNet stages.}
  \label{main_comparison}
\end{figure*}

Many methods have been proposed for multi-scale feature extraction and fusion~\citep{FPN,PAFPN,NAS-FPN,BiFPN}. One widely-used model is the feature pyramid network (FPN)~\citep{FPN} (Figure~\ref{main_comparison} (a)). FPN consists of a pre-designed classification backbone for extracting multi-scale features and a lightweight fusion module for fusing these features. Moving beyond the FPN, some cascade fusion strategies have been developed and showcased efficacy in multi-scale feature fusion~\citep{PAFPN,NAS-FPN,BiFPN}. \mbox{Figure~\ref{main_comparison} (b)} shows the structure of the representative BiFPN~\citep{BiFPN}. It iteratively fuses multi-scale features using repeated bottom-up and top-down pathways. However, the time for feature fusion in these networks is relatively late, because they allocate most computational resources to the classification backbone to extract the initial multi-scale features. We define \textit{the time for feature fusion} as the ratio of the parameters of the sub-network before the first fusion module to the whole \mbox{network}. A smaller ratio indicates an earlier time. For instance, considering an FPN built based on the classification model ConvNeXt-S~\citep{convnext}, the time for feature fusion is the ratio of the parameters of the backbone ConvNeXt-S to the entire FPN (91.7\%). Given the complexity of dense prediction tasks where models are required to handle objects of diverse sizes, we expect that integrating early multi-scale feature fusion within the backbone could enhance model performance.

Some methods have transitioned from using pre-designed classification networks to designing task-specific backbones for dense prediction tasks~\citep{HRNet,SpineNet,GiraffeDet,RovCol}. In these methods, some incorporate early multi-scale feature fusion. For example, HRNet~\citep{HRNet}, one of the representative works (Figure~\ref{main_comparison} (c)), aims to learn semantically rich and spatially precise features. Although HRNet performs the first feature fusion very early (2.7\%), it generates high-level (low-resolution) features with strong semantic information quite late. This limits their role in guiding the learning of low-level (high-resolution) features that are important for dense prediction tasks. In contrast, SpineNet~\citep{SpineNet} employs neural architecture search (NAS)~\citep{NAS} to learn a scale-permuted backbone with early feature fusion. Nevertheless, the resulting network is complex and exhibits limited performance when transferred to different detectors~\citep{SpineNet}. GiraffeDet~\citep{GiraffeDet} integrates a lightweight backbone with a heavy fusion module for object detection, aiming to enhance the information exchange between high-level and low-level features. Yet, it fuses multi-scale features in a fully connected way, which inevitably increases runtime latency. A detailed discussion of related works can be found in Section~\ref{related_work}. Clearly, an appropriate structure for effective early multi-scale feature fusion is lacking.

In this paper, we present CEDNet, a cascade encoder-decoder network tailored for dense prediction tasks. CEDNet begins with a stem module to extract initial high-resolution features. Following this, CEDNet incorporates several cascade stages to generate multi-scale features, with all stages sharing the same encoder-decoder structure. The encoder-decoder structure can be realized in various ways. Figure~\ref{main_comparison} (d) illustrates a three-stage CEDNet built on the FPN-style design. CEDNet evenly allocates its computational resources across stages and fuses multi-scale features within each decoder. As a result, CEDNet performs multi-scale feature fusion from the early stages of the network. This strategy ensures that high-level features from the early stages are integrated to guide the \mbox{learning of} low-level features in subsequent stages. Moreover, CEDNet possesses a more streamlined and efficient structure, making it suitable for a wide variety of models and tasks.

We investigated three well-known methods, \ie Hourglass~\citep{Hourglass}, UNet~\citep{UNet}, and FPN~\citep{NAS-FPN}, as the encoder-decoder structure in experiments and found that they all performed well. Due to the slightly better results of the FPN, it is adopted as the default encoder-decoder structure in CEDNet for further analysis on object detection, instance segmentation and semantic segmentation. On the COCO \textit{val 2017} for object detection and instance segmentation, the CEDNet variants outperformed their counterparts, \ie the ConvNeXt variants~\citep{convnext}, achieving an increase of 1.9-2.9 \% in box AP and 1.2-1.8\% in mask AP based on the popular framework RetinaNet~\citep{ResNet} and Mask R-CNN~\citep{MaskRCNN}. On the ADE20k for semantic segmentation, the CEDNet variants outperformed their counterparts by 0.8-2.2\% mIoU based on the renowned framework UperNet~\citep{UPerNet}. These results demonstrate the excellent performance of CEDNet and encourage the community to rethink the prevalent model design principle for dense prediction tasks.

\section{Related work}

\subsection{Multi-scale feature fusion}
Many methods adopt pre-designed classification backbones to extract multi-scale features. However, the low-level features produced by traditional classification networks are semantically weak and ill-suited for downstream dense prediction tasks. To tackle this limitation, many strategies~\citep{FPN,PAFPN,BiFPN,Deeplabv3,NAS-FPN} have been proposed for multi-scale feature fusion. In semantic segmentation, DeeplabV3+~\citep{Deeplabv3} fuses low-level features with semantically strong high-level features produced by atrous spatial pyramid pooling. In object detection, FPN~\citep{FPN} introduces a top-down pathway to sequentially combine high-level features with low-level features. NAS-FPN~\citep{NAS-FPN} fuses multi-scale features by repeated fusion stages searched by neural architecture search~\citep{NAS}. EfficientDet~\citep{BiFPN} adopts a weighted bi-directional feature pyramid network in conjunction with a compound scaling rule to achieve efficient feature fusion. A common drawback of these methods is that they allocate most computational resources to the classification backbone, delaying feature fusion and potentially undermining fusion effectiveness. In contrast, our approach evenly allocates computational resources to multiple stages and perform feature fusion within each stage.

\subsection{Backbone designs for dense prediction}\label{related_work}
Instead of fusing multi-scale features from pre-designed classification networks, some studies have attempted to design task-specific backbones for dense prediction tasks~\citep{UNet,Hourglass,HRNet,SpineNet,GiraffeDet,CBNet,DetectoRS,RovCol}. For instance, UNet~\citep{UNet} employs a U-shape structure to acquire high-resolution and semantically strong features in medical image segmentation. Hourglass~\citep{Hourglass} introduces a convolutional network consisting of repeated bottom-up and top-down pathways for human pose estimation. HRNet~\citep{HRNet} retains high-resolution features throughout the whole network and performs well in semantic segmentation and human pose estimation. In object detection, SpineNet~\citep{SpineNet} leverages neural architecture search~\citep{NAS} to learn scale-permuted backbones. GiraffeDet~\citep{GiraffeDet} pairs a lightweight backbone with a heavy fusion module to encourage dense information exchange among multi-scale features. RevCol~\citep{RovCol} feeds the input image to several identical subnetworks simultaneously and connects them through reversible transformations.

While the aforementioned methods incorporate early multi-scale feature fusion, they either exhibit effectiveness solely on specific models and tasks~\citep{UNet,Hourglass,SpineNet}, or do not fully harness the potential of high-level features to guide the learning of low-level features~\citep{HRNet,RovCol}. In contrast, our method incorporates multiple cascade stages to iteratively extract and fuse multi-scale features. Therefore, the high-level features from early stages can be integrated to instruct the learning of low-level features in subsequent stages. Moreover, CEDNet showcases excellent performance across a broad spectrum of models and tasks.

\section{CEDNet}

\subsection{Overall architecture}

\begin{figure*}[t]
  \centering
  \hspace*{-3mm}
  \includegraphics[width=0.97\textwidth]{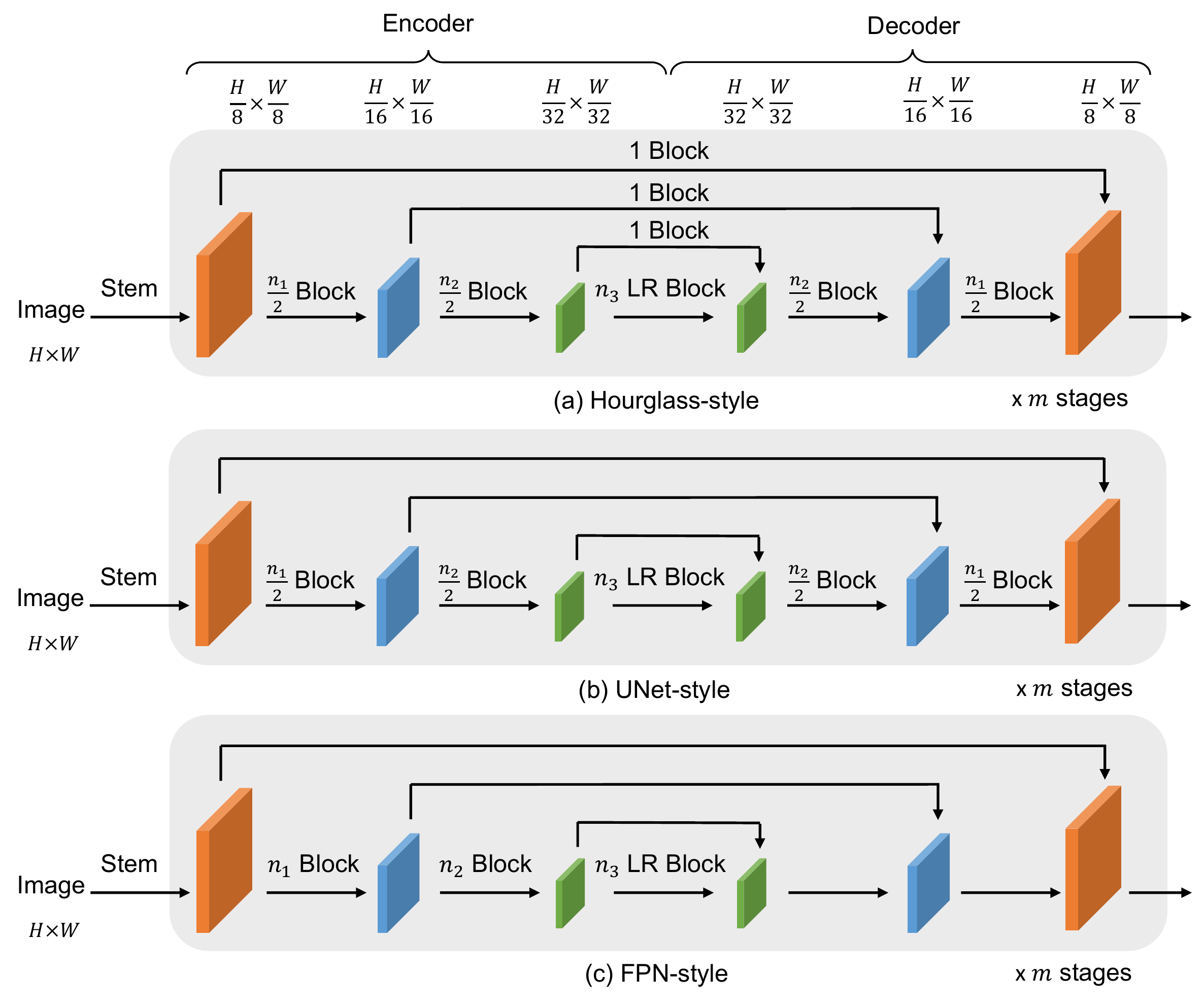}
  \vspace*{-4mm}
  \caption{Three implementations of CEDNet. The input image with a spatial size of $H$$\times $$W$ is fed into a lightweight stem module to extract high-resolution features of size $\frac{H}{8}$$\times $$\frac{W}{8}$. These features are then processed through $m$ cascade stages to extract multi-scale features. Block denotes CED block, and LR Block denotes long-range (LR) CED block. The down-sampling layers (2$\times$2 convolution with stride 2) and the up-sampling layers (bilinear interpolation) are omitted for clarity.}
  \label{architecture}
  \vspace*{-2mm}
\end{figure*}

Figure~\ref{architecture} illustrates the overall architecture of CEDNet. The input RGB image with a spatial size of $H$$\times$$W$ is fed into a stem module to extract high-resolution feature maps of size $\frac{H}{8}$$\times $$\frac{W}{8}$. The stem module comprises two sequential 3$\times$3 convolutional layers (each with a stride of 2), $n_0$ CED blocks, and a 2$\times$2 convolutional layer with a stride of 2. Each 3$\times$3 convolutional layer is followed by a LayerNorm~\citep{LN} layer and a GELU~\citep{GELU} unit. The further details about the CED block can be found in Section~\ref{block_designs}. Subsequently, $m$ cascade stages, each with the same encoder-decoder structure, are utilized to extract multi-scale features. The multi-scale features from the final decoder are then fed into downstream dense prediction tasks. \textit{Unlike in FPN and BiFPN, no extra feature fusion modules are required after the CEDNet backbone.} We discuss three implementations of the encoder-decoder structure in Section~\ref{three_implementations}.

\subsection{Three encoder-decoder structures}\label{three_implementations}

In CEDNet, each stage employs an encoder-decoder structure. The encoder extracts multi-scale features, while the decoder integrates these features into single-scale, highest-resolution ones. Consequently, the high-level (low-resolution) features from early stages are integrated to guide the learning of low-level features in subsequent stages. While many methods can be used to realize the encoder-decoder structure, we adopt three well-known methods for our purposes in this study.

\noindent\textbf{Hourglass-style.} The Hourglass network~\citep{Hourglass} is a deep learning architecture specifically designed for human pose estimation. It resembles an encoder-decoder design but stands out with its symmetrical hourglass shape, from which its name is derived. In this study, we draw inspiration from the Hourglass architecture to devise an hourglass-style encoder-decoder (Figure~\ref{architecture} (a)). In alignment with the original design, a CED block is employed to transform the feature maps from the encoder before integrating them into the symmetrical feature maps in the decoder.

\noindent\textbf{UNet-style.} UNet is a prominent network primarily employed in medical image segmentation~\citep{UNet}. Recent advancements have also shown its successful application in diffusion models~\citep{controlnet}. As illustrated in Figure~\ref{architecture} (b), the UNet-style encoder-decoder has a symmetrical shape. Unlike the hourglass-style design, identity skip connections are harnessed to bridge the symmetrical feature maps between the encoder and the decoder.

\noindent\textbf{FPN-style.} FPN~\citep{FPN} is initially designed for object detection and instance segmentation, aiming to fuse multi-scale features from pre-designed classification networks. In this work, we incorporate the FPN-style encoder-decoder as a separate stage in CEDNet, as shown in Figure~\ref{architecture} (c). Different from the standard FPN implementation, we eliminate the 3$\times$3 convolutions responsible for transforming the merged symmetrical feature maps. As a result, most computational resources are allocated to the encoders, with only two 1$\times$1 convolutions in each decoder for feature channel alignment.

\subsection{Block designs}\label{block_designs}

\begin{minipage}[b]{0.73\linewidth}
  \textbf{CED block.} \textit{The solid elements} in Figure~\ref{ced_block} illustrate the general structure of the CED block. This block comprises a token mixer for spatial feature interactions and a multi-layer perceptron (MLP) with two layers for channel feature interactions. The token mixer can be various existing designs, such as the 3$\times$3 convolution in ResNet~\citep{ResNet}, the 7$\times$7 depth-wise convolution in ConvNeXt~\citep{convnext}, and the local window attention in Swin transformer~\citep{swin}. In CEDNet, we take the lightweight 7$\times$7 depth-wise convolution from ConvNeXt as the default token mixer. \textit{Please note that a more powerful token mixer may yield enhanced performance, but that is not the focus of this work.}

  \vspace*{2mm}
  \textbf{LR CED block.} To increase the receptive field of neurons, we introduce the LR CED block. Beyond the CED block, this block incorporates a 7$\times$7 \textit{dilated} depth-wise convolution accompanied by two skip connections, as highlighted by the dashed elements. By integrating the dilated depth-wise convolution, the LR CED block is capable of capturing long-range dependencies among spatial features with only a marginal increase in parameters and computational overhead. The LR CED blocks are utilized to transform the lowest-resolution features in each CEDNet stage.
\end{minipage}\hspace*{1mm}
\begin{minipage}[b]{0.26\linewidth}
  \centering
  \includegraphics[width=0.9\textwidth]{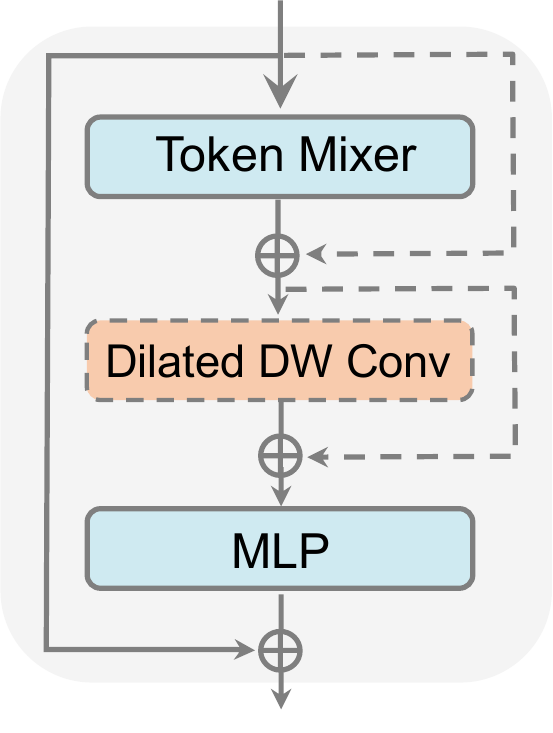}
  \vspace*{-4mm}
  \captionof{figure}{\small Structure of the CED block and the LR CED block. The CED block consists solely of solid elements, while the LR CED block includes both solid and dashed elements. DW Conv is short for depth-wise convolution.}
  \label{ced_block}
\end{minipage}

\subsection{Architecture specifiactions}\label{arch_specifiactions}
We have constructed three CEDNet variants based on the FPN-style encoder-decoder, \ie CEDNet-NeXt-T/S/B, where the suffixes T/S/B indicate the model size tiny/small/base. We take the 7$\times$7 depth-wise convolution from ConvNeXt as the default token mixer for all (LR) CED blocks. For all LR CED blocks, we set the dilation rate $r$ of the dilated convolution to 3. These variants adopt different channel dimensions $C$, different numbers of blocks $B$ = ($n_0$, $n_1$, $n_2$, $n_3$), and different numbers of stages $m$. The configuration hyper-parameters for these variants are presented below:
\begin{itemize}
  \item CEDNet-NeXt-T: $C$=(96, 192, 352, 512), $B$=(3, 2, 4, 2), $m$=3
  \item CEDNet-NeXt-S: $C$=(96, 192, 352, 512), $B$=(3, 2, 7, 2), $m$=4
  \item CEDNet-NeXt-B: $C$=(128, 256, 448, 704), $B$=(3, 2, 7, 2), $m$=4
\end{itemize}

\section{Experiments}

\subsection{Three encoder-decoder structures}\label{encoder_decoder_exps}
We conducted experiments to compare the three encoder-decoder structures.

\noindent\textbf{Pre-training settings.} Following common practice~\citep{swin,convnext}, we pre-trained CEDNet on the ImageNet-1K dataset~\citep{ImageNet}. The ImageNet-1K dataset consists of 1000 object classes with 1.2M training images. To perform classification, we removed the last decoder and attached a classification head on the lowest-resolution features from the last stage. The hyperparameters, augmentation and regularization strategies strictly follows~\citep{convnext}.

\noindent\textbf{Pre-training results.} We built the CEDNet models using three different encoder-decoder structures.  Appendix A presents the results of the CEDNet models on ImageNet-1K in comparison with some recent methods in image classification. We report the top-1 accuracy \mbox{on the} validation set. Table A1 shows that the CEDNet models slightly outperformed their counterparts, \ie the ConvNeXt variants. \textit{Please note that the CEDNet is specifically designed for dense prediction tasks, and surpassing state-of-the-art methods in image classification is not our goal.}

\noindent\textbf{Fine-tuning settings.} We fine-tuned models on object detection with COCO 2017~\citep{COCO} based on the well-known detection framework RetinaNet~\citep{RetinaNet} using the MMDetection toolboxes~\citep{mmdet}. For training settings, we mainly followed~\citep{convnext}. Additionally, we found that the proposed CEDNet models were easy to overfit the training data. To fully explore the potential of CEDNet models, we used large scale jittering and copy-and-paste data augmentation following~\citep{CopyPaste}, but only with box annotations. We re-trained all baseline models with the same data augmentation for a fair comparison.

\begin{table}[t]
    \caption{Comparison among three encoder-decoder structures. $AP^b$ is the overall detection accuracy on COCO \textit{val2017}. The model inference speed FPS was measured on a single RTX 3090 GPU.} 
    \label{results_encoder_decoder}
    \centering
    \setlength{\tabcolsep}{1mm}{}
    \scalebox{0.97}{
    \begin{tabular}{l|cc|cccc}
    \toprule
    Method & Param & FLOPs & AP$^{b}$ & AP$_{50}^b$ & AP$_{75}^b$ & FPS$\uparrow$\\
    \midrule
    ConvNeXt-T w/ FPN~\citep{FPN}  & 39M & 243G & 45.4 & 66.5 & 48.7 & 20.0 \\
    ConvNeXt-T w/ NAS-FPN~\citep{NAS-FPN} & 47M & 289G & 46.6 & 66.8 & 49.9 & 18.1 \\
    ConvNeXt-T w/ BiFPN~\cite{BiFPN} & 39M & 248G & 46.7 & 67.1 & 50.2 & 16.8 \\
    \midrule
    HRNet-w32~\citep{HRNet} & 39M & 320G & 45.7 & 66.5 & 49.2 & 15.9 \\
    SpineNet-96~\citep{SpineNet} & 43M & 265G & 47.1 & 67.1 & 51.1 & 16.3 \\
    GiraffeDet-D11~\citep{GiraffeDet} & 69M & 275G & 46.6 & 65.0 & 51.1 & 12.4 \\
    \midrule
    CEDNet-NeXt-T (Hourglass-style) & 39M & 255M & 47.4 & 68.5 & 50.7 & 15.9 \\
    CEDNet-NeXt-T (UNet-style)& 39M & 255M & 47.9 & 68.9 & 51.4 & 16.7 \\
    CEDNet-NeXt-T (FPN-style) & 39M & 255M & \textbf{48.3} & \textbf{69.1} & \textbf{51.6} & 17.1 \\
    \bottomrule
    \end{tabular}
    }
\end{table}

\noindent\textbf{Fine-tuning results.} Table~\ref{results_encoder_decoder} shows that all three CEDNet models yielded significant gains over the models with FPN, NAS-FPN, and BiFPN, all of which utilize the classification network ConvNeXt-T to extract initial multi-scale features. This result validates the effectiveness of the cascade encoder-decoder network that performs multi-scale feature fusion from early stages. In addition, the CEDNet models surpassed other early feature fusion methods: HRNet, SpineNet, and GiraffeDet. We attempted to pre-train the entire BiFPN model by attaching a classification head to the coarsest feature maps of the last bottom-up pathway in the fusion module. However, we obtained poor results, \ie 76.8\% top-1 accuracy on ImageNet and 39.4\% box AP on COCO.

Since the model built on the FPN-style encoder-decoder slightly outperformed the models built on the UNet-style and Hourglass-style encoder-decoder in both detection accuracy and model inference speed, \textit{we adopted the FPN-style encoder-decoder for CEDNet by default in subsequent experiments.}

\subsection{Object detection on COCO}\label{detection_exps}

\begin{table}[t]
  \caption{Results of object detection and instance segmentation on the COCO \textit{val2017}. AP$^b$ and AP$^m$ are the overall metrics for object detection and instance segmentation, respectively. If required, FPN was adopted as the default fusion module for methods except CEDNet and the models marked by $^\dag$} 
  \label{results_coco_val2017}
   \centering
   \setlength{\tabcolsep}{1.2mm}{}
   \scalebox{0.97}{
      \begin{tabular}{l|cc|cccccc}
      \toprule
      \multicolumn{9}{c}{(a) Deformable DETR} \\
      \midrule
      Method  & Param & FLOPs &
      \color{black}AP$^b$ & AP$^b_{50}$ & AP$^b_{75}$ & AP$^b_{\rm{S}}$ & AP$^b_{\rm{M}}$ & AP$^b_{\rm{L}}$ \\
       \midrule
      ConvNeXt-T & 42M & 231G & 47.1 & 66.7 & 51.8 & 28.8 & 50.5 & 62.3 \\
      ConvNeXt-S & 64M & 317G & 49.0 & 68.6 & 53.7 & 31.2 & 52.4 & 64.5 \\
      \rowcolor{gray! 20} CEDNet-NeXt-T & 43M & 223G & 49.3 & 69.1 & 53.7 & 32.1 & 52.8 & 65.3 \\
      \rowcolor{gray! 20} CEDNet-NeXt-S & 65M & 304G & 50.3 & 70.2 & 55.2 & 32.3 & 54.6 & 65.2 \\
      \midrule
      \multicolumn{9}{c}{(b) RetinaNet} \\
      \midrule
      Backbone  & Param & FLOPs &
      \color{black}AP$^b$ & AP$^b_{50}$ & AP$^b_{75}$ & AP$^b_{\rm{S}}$ & AP$^b_{\rm{M}}$ & AP$^b_{\rm{L}}$ \\
      \midrule
      SpineNet-143$^\dag$~\citep{SpineNet} & 67M & 524G & 48.1 & 67.6 & 52.0 & 30.2 & 51.1 & 59.9 \\
      Swin-T~\citep{swin} & 39M & 245G & 45.0 & 65.9 & 48.4 & 29.7 & 48.9 & 58.1 \\
      Swin-S~\citep{swin} & 60M & 335G & 46.4 & 67.0 & 50.1 & 31.0 & 50.1 & 60.3 \\
      Swin-B~\citep{swin} & 98M & 477G & 45.8 & 66.4 & 49.1 & 29.9 & 49.4 & 60.3 \\
      \midrule
      ConvNeXt-T & 39M & 243G & 45.4 & 67.0 & 48.7 & 29.5 & 49.9 & 59.9 \\
      ConvNeXt-S & 60M & 329G  & 47.4 & 68.3 & 51.2 & 32.0 & 51.5 & 61.6 \\
      \rowcolor{gray! 20} CEDNet-NeXt-T & 39M & 255G & \color{black!100}48.3 & 69.1 & 51.6 & 33.2 & 53.1 & 62.7 \\
      \rowcolor{gray! 20} CEDNet-NeXt-S & 61M & 335G & \color{black!100}49.6 & 70.8 & 53.2 & 34.8 & 54.0 & 63.5 \\
      \midrule
      \multicolumn{9}{c}{(c) Mask R-CNN} \\
      \midrule
      Backbone  & Param & FLOPs &
      \color{black}AP$^b$ & AP$^b_{50}$ & AP$^b_{75}$ & \color{black}AP$^m$ & AP$^m_{50}$ & AP$^m_{75}$ \\
      \midrule
      DetectoRS-50$^\dag$~\citep{DetectoRS} & 105M & 432G & 46.2 & 65.1 & 50.2 & 40.4 & 62.5 & 43.5 \\
      Swin-T~\citep{swin} & 48M & 264G & 46.0 & 68.1 & 50.3 & 41.6 & 65.1 & 44.9 \\
      Swin-S~\citep{swin} & 69M & 354G & 48.5 & 70.2 & 53.5 & 43.3 & 67.3 & 46.6 \\
      FocalNet-S~\citep{FocalNet} & 72M & 365G & 49.3 & 50.9 & 54.6 & 44.1 & 67.9 & 47.4 \\
      Swin-B~\citep{swin} & 107M & 496G & 48.5 & 69.8 & 53.2 & 43.4 & 66.8 & 46.9 \\
      FocalNet-B~\citep{FocalNet} & 114M & 507G & 49.8 & 70.7 & 54.2 & 43.8 & 68.2 & 47.2 \\
      \midrule
      ConvNeXt-T & 48M & 262G & 46.4 & 68.1 & 51.3 & 42.3 & 65.2 & 45.9 \\
      ConvNeXt-S & 70M & 348G  & 48.5 & 70.0 & 53.3 & 43.8 & 67.2 & 47.7 \\
      \rowcolor{gray! 20} CEDNet-NeXt-T & 49M & 274G & \color{black!100}49.2 & 70.3 & 53.7 & \color{black!100}44.1 & 67.8 & 47.5 \\
      \rowcolor{gray! 20} CEDNet-NeXt-S & 72M & 355G & \color{black!100}50.4 & 71.7 & 55.1 & \color{black!100}45.0 & 68.9 & 48.6 \\
      \midrule
      \multicolumn{9}{c}{(d) Cascade Mask R-CNN} \\
      \midrule
      Backbone  & Param & FLOPs &
      \color{black}AP$^b$ & AP$^b_{50}$ & AP$^b_{75}$ & \color{black}AP$^m$ & AP$^m_{50}$ & AP$^m_{75}$ \\
      \midrule
      CBNet-X152~\citep{CBNet} & 238M & 1358G & 50.7 & 69.8 & 55.5 & 43.3 & 66.9 & 46.8 \\
      Swin-T~\citep{swin} & 86M & 745G & 50.5 & 69.3 & 54.9 & 43.7 & 66.6 & 47.1 \\
      RovCol-T~\citep{RovCol} & 88M & 741G & 50.6 & 68.9 & 54.9 & 43.8 & 66.7 & 47.4 \\
      Swin-S~\citep{swin} & 107M & 838G & 51.8 & 70.4 & 56.3 & 44.7 & 67.9 & 48.5 \\
      RovCol-S~\citep{RovCol} & 118M & 833G & 52.6 & 71.1 & 56.8 & 45.5 & 68.8 & 49.0 \\
      Swin-B~\citep{swin} & 145M & 982G & 51.9 & 70.5 & 56.4 & 45.0 & 68.1 & 48.9 \\
      RovCol-B~\citep{RovCol} & 196M & 988G & 53.0 & 71.4 & 57.3 & 45.9 & 69.1 & 50.1 \\
      \midrule
      ConvNeXt-T & 86M & 741G & 50.8 & 69.4 & 55.2 & 44.5 & 66.9 & 48.5 \\
      ConvNeXt-S & 108M & 827G  & 51.9 & 71.0 & 56.6 & 45.4 & 68.6 & 49.5 \\
      ConvNeXt-B & 146M & 964G & 52.7 & 71.3 & 57.2 & 45.6 & 68.9 & 49.5 \\
      \rowcolor{gray! 20} CEDNet-NeXt-T & 87M & 753G & \color{black!100}52.5 & 71.4 & 56.8 & \color{black!100}45.9 & 69.0 & 49.7  \\
      \rowcolor{gray! 20} CEDNet-NeXt-S & 110M & 833G & \color{black!100}53.5 & 72.4 & 58.1 & \color{black!100}46.7 & 69.9 & 50.6 \\
      \rowcolor{gray! 20} CEDNet-NeXt-B & 148M & 968G & \color{black!100}53.6 & 72.6 & 57.8 & \color{black!100}46.9 & 70.2 & 51.0 \\
      \bottomrule
      \end{tabular}
   }
\end{table}

\noindent\textbf{Settings.} We benchmark our models on object detection with COCO 2017~\citep{COCO} based on four representative frameworks, \ie Deformable DETR~\citep{defromable_detr}, RetinaNet~\citep{RetinaNet}, Mask R-CNN~\citep{MaskRCNN}, and Cascade Mask R-CNN~\citep{Cascade-RCNN}. All training settings were same as the fine-tuning settings in Section~\ref{encoder_decoder_exps}.

\noindent\textbf{Main results.} Table~\ref{results_coco_val2017} presents the object detection results of the CEDNet models to compare with other methods. The CEDNet models yielded significant gains over the ConvNeXt models. Specifically, CEDNet-NeXt-T achieved 2.2\%, 2.9\%, 2.8\%, and 1.7\% box AP improvements over its counterpart ConvNeXt-T based on the Deformable DETR, RetinaNet, Mask R-CNN, and Cascade Mask R-CNN, respectively. When scaled up to CEDNet-NeXt-S, CEDNet still outperformed its baseline ConvNeXt-S by 1.3\%, 2.2\%, 1.9\%, and 1.6\% box AP based on the four detectors.

\begin{table}[t]
  \caption{Results of semantic segmentation on the ADE20K \textit{validation} set.  The superscripts $^{ss}$ and $^{ms}$ denote single-scale and multi-scale testing. FPN was adopted as the default fusion module for methods except CEDNet. No extra fusion modules were required after the CEDNet backbone.}
  \label{results_ade20k}
   \centering
   \setlength{\tabcolsep}{1.2mm}{}
   \scalebox{0.97}{
      \begin{tabular}{l|cc|c|cc}
      \toprule
      Method & Param. & FLOPs & Input size & mIoU$^{ss}$ & mIoU$^{ms}$ \\
      \midrule
      Focal-T~\citep{FocalTransformer} & 62M & 998G & 512$^2$ & 45.5 & 47.0 \\
      RovCol-T~\citep{RovCol} & 60M & 937G & 512$^2$ & 47.4 & 47.6 \\
      Swin-S~\citep{swin} & 81M & 1038G & 512$^2$ & 47.6 & 49.5 \\
      Focal-S~\citep{FocalTransformer} & 85M & 1130G & 512$^2$ & 48.0 & 50.0 \\
      RovCol-S~\citep{RovCol} & 90M & 1031G & 512$^2$ & 47.9 & 49.0 \\
      Swin-B~\citep{swin} & 121M & 1188G & 512$^2$ & 48.1 & 49.7 \\
      Focal-B~\citep{FocalTransformer} & 126M & 1354G & 512$^2$ & 49.0 & 50.5 \\
      RovCol-B~\citep{RovCol} & 122M & 1169G & 512$^2$ & 49.0 & 50.1 \\
      \midrule
      ConvNeXt-T~\citep{convnext} & 60M & 939G & 512$^2$ & 46.0 & 46.7 \\
      ConvNeXt-S~\citep{convnext} & 82M & 1027G & 512$^2$ & 48.7 & 49.6 \\
      ConvNeXt-B~\citep{convnext} & 122M & 1170G & 512$^2$ & 49.1 & 49.9 \\
      \rowcolor{gray! 20} CEDNet-NeXt-T & 61M & 962G & 512$^2$ & 48.3 & 48.9 \\
      \rowcolor{gray! 20} CEDNet-NeXt-S & 83M & 1045G & 512$^2$ & 49.8 & 50.4 \\
      \rowcolor{gray! 20} CEDNet-NeXt-B & 123M & 1184G & 512$^2$ & 49.9 & 51.0 \\
      \bottomrule
      \end{tabular}
   }
\end{table}

\subsection{Instance segmentation on COCO}

\noindent\textbf{Settings.} We conducted experiments on instance segmentation with COCO 2017~\citep{COCO} based on the commonly used Mask R-CNN~\citep{MaskRCNN} and Cascade Mask R-CNN~\citep{Cascade-RCNN} following~\citep{convnext,swin}. These two frameworks perform object detection and instance segmentation in a multi-task manner. All training settings were same as Section~\ref{encoder_decoder_exps}.

\noindent\textbf{Main results.} Table~\ref{results_coco_val2017} presents the instance segmentation results (see the columns for metrics AP$^m$, AP$^m_{50}$, and AP$^m_{75}$). Based on Mask R-CNN, the models CEDNet-NeXt-T and CEDNet-NeXt-S outperformed their counterparts ConvNeXt-T and ConvNeXt-S by 1.8\% and 1.2\% mask AP, respectively. When applied to the more powerful Cascade Mask R-CNN, the proposed CEDNet models still  yielded 1.3-1.4\% mask AP gains over the baseline models. These improvements were consistent with those in object detection. When scaled up to the larger model CEDNet-NeXt-B, CEDNet achieved 46.9\% mask AP based on the Cascade Mask R-CNN.

\subsection{Semantic segmentation on ADE20k}
\noindent\textbf{Settings.} We conducted experiments on semantic segmentation with the ADE20k~\citep{ADE20K} dataset based on UperNet~\citep{UPerNet} using the MMSegmentation~\citep{mmseg} toolboxes, and report the results on the validation set. The training settings strictly follow~\citep{convnext}. As the data augmentation strategies used for semantic segmentation were strong enough to train the proposed CEDNet models, no extra data augmentation was introduced. 

\noindent\textbf{Main results.} Table~\ref{results_ade20k} presents the semantic segmentation results. Compared with the ConvNeXt models, CEDNet achieved 0.8-2.2\% mIoU gains in the multi-scale test setting with different model variants, which demonstrates the effectiveness of our method in semantic segmentation.

\subsection{Ablation studies}\label{ablation_studies}
To better understand CEDNet, we ablated some key components and evaluated the performance in object detection based on CEDNet-NeXt-T and RetinaNet. Models in Tables \ref{motivation_exp} and \ref{ablation_number_of_stages} were pre-trained on ImageNet for 100 epochs and fine-tuned on COCO for 12 epochs. The other models were trained under the same settings as Section~\ref{encoder_decoder_exps}.

\begin{table}[t]
  \begin{minipage}[b]{0.48\linewidth}
    \caption{Early feature fusion. $n_1^i$, $n_2^i$, $n_3^i$ are the number of blocks in the $i$-th stage.}
    \label{motivation_exp}
    \centering
    \setlength{\tabcolsep}{1mm}{}
    \scalebox{0.91}{
      \begin{tabular}{c|c|c|c|cc}
        \toprule
        Time & \#Stage & $n_1^1$, $n_2^1$, $n_3^1$  & $n_1^2$, $n_2^2$, $n_3^2$ & Param &
        AP$^b$ \\
        \midrule
        6/6 & 2 & 6, 9, 3 & -       & 38M & 40.6            \\[0.4mm]
        5/6 & 2 & 5, 10,5 & 1, 2, 1 & 38M & 42.2            \\[0.4mm]
        4/6 & 2 & 4, 8, 4 & 2, 4, 2 & 38M & 42.4            \\[0.4mm]
        3/6 & 2 & 3, 6, 3 & 3, 6, 3 & 38M & 42.9            \\[0.4mm]
        2/6 & 2 & 2, 4, 2 & 4, 8, 4 & 38M & \textbf{43.3}   \\[0.4mm]
        1/6 & 2 & 1, 2, 1 & 5,10, 5 & 38M & \textbf{43.3}   \\
        \midrule
        2/6 & 3 & 2, 4, 2 & 2, 4, 2 & 39M & \textbf{43.3}   \\
        \bottomrule
      \end{tabular}
      }
  \end{minipage}\hspace*{2mm}
  \begin{minipage}[b]{0.48\linewidth}
    \caption{Different token mixers. WA and DW are short for window attention and depth-wise.}
    \label{different_blocks}
    \centering
    \setlength{\tabcolsep}{1mm}{}
    \scalebox{0.8}{
      \begin{tabular}{l|c|cc}
        \toprule
        Backbone & Token mixer & Param & AP$^b$ \\
        \midrule
        ResNet-50 & \multirow{2}{*}{Vanilla conv 3$\times$3} & 38M & 41.7 \\
        CEDNet-R50-T & & 39M & \textbf{45.3}\\
        \midrule
        Swin-T & \multirow{2}{*}{Local WA} & 39M & 44.9 \\
        CEDNet-Swin-T & & 37M & \textbf{47.4} \\
        \midrule
        ConvNeXt-T & \multirow{2}{*}{DW conv 7$\times$7} & 39M & 45.4 \\
        CEDNet-NeXt-T & & 39M & \textbf{48.3} \\
        \midrule
        CSwin-T & \multirow{2}{*}{Cross WA} & 32M & 48.0 \\
        CEDNet-CSwin-T & & 33M & \textbf{49.5} \\
        \bottomrule
      \end{tabular}
    }
  \end{minipage}

  \vspace*{4mm}
  \begin{minipage}[b]{0.33\linewidth}
    \caption{Number of stages.}
    \label{ablation_number_of_stages}
    \centering
    \setlength{\tabcolsep}{1mm}{}
    \scalebox{0.9}{
      \begin{tabular}{c|c|cc}
      \toprule
        $m$ & $n_1$, $n_2$, $n_3$ & Param & AP$^b$ \\
      \midrule
        1 & 6, 9, 3 & 38M & 40.6   \\
        2 & 3, 6, 3 & 38M & 42.9   \\
        3 & 2, 4, 2 & 39M & \textbf{43.3}   \\
        4 & 1, 4, 1 & 39M & 43.1   \\
      \bottomrule
      \end{tabular}
    }
  \end{minipage}
  \begin{minipage}[b]{0.33\linewidth}
    \caption{Data augmentation.}
    \label{ablation_stronger_aug}
    \centering
      \setlength{\tabcolsep}{1mm}{}
      \scalebox{0.9}{
      \begin{tabular}{l|c|l}
        \toprule
        Backbone & Aug. & AP$^b$ \\
        \midrule
        ConvNeXt-T & Existing & 45.2   \\
        ConvNeXt-T & Ours & 45.4   \\
        CEDNet-NeXt-T & Existing & 47.0    \\
        CEDNet-NeXt-T & Ours & \textbf{48.3}   \\
        \bottomrule
      \end{tabular}
      }
  \end{minipage}
  \begin{minipage}[b]{0.33\linewidth}
    \caption{LR CED block.}
    \label{ablation_focal_block}
    \centering
      \setlength{\tabcolsep}{1mm}{}
      \scalebox{0.9}{
      \begin{tabular}{c|cc}
      \toprule
      LR block & Param & AP$^b$ \\
      \midrule
                  & 38.5M & 47.9   \\
      \checkmark & 38.6M & \textbf{48.3}   \\
      \bottomrule
      \end{tabular}
  }
  \end{minipage}

\end{table}

  \noindent\textbf{Effectiveness of early feature fusion.} To explore the effectiveness of early feature fusion, we constructed several two-stage CEDNet-NeXt-T models varying in fusion time. We modulated the fusion time of each model by adjusting the computational resources allocated to each stage. All models have a similar size. Table~\ref{motivation_exp} shows that the detection accuracy (AP$^b$) gradually improved as the time for multi-scale feature fusion becomes earlier, which demonstrates that early feature fusion is beneficial for dense prediction tasks. Although the two-stage CEDNet models which allocate a proper proportion of computational resources to each stage performed well, we adopted the same configuration for all stages by default to simplify the structure design and employed three stages to achieve early feature fusion instead (the last row in Table~\ref{motivation_exp}). 

\noindent\textbf{Effectiveness on different token mixers.} We constructed the CEDNet models with various token mixers and compared the resulting models with their counterparts (Table~\ref{different_blocks}). The CEDNet models consistently surpassed their counterparts by 1.5-3.6\% mAP, which underscores the generality of our CEDNet. \textit{While CEDNet yielded better performance when utilizing the more powerful cross-window attention introduced in CSwin Transformer~\citep{CSWin}, we opted for the more representative ConvNeXt as our baseline in this work and took the 7$\times$7 depth-wise convolution from ConvNeXt as the default token mixer for CEDNet.}

\noindent\textbf{Different numbers of stages.} We built the CEDNet-NeXt-T models with different numbers of stages while maintaining the same configurations across stages. A model with more stages performs multi-scale feature fusion earlier. Table~\ref{ablation_number_of_stages} shows that the CEDNet-NeXt-T model with three stages achieved the best detection accuracy. Intuitively, a model with more stages can fuse features more sufficiently, but more network connections may make it harder to optimize.

\noindent\textbf{Influence of data augmentation.} We compared the data augmentation strategy used by~\citep{convnext} and the enhanced data augmentation strategy we adopted for detection fine-tuning. Table~\ref{ablation_stronger_aug} shows that the ConvNeXt models achieved similar results under both settings, but our CEDNet model exhibited notable improvements with the enhanced data augmentation. This may be because that CEDNet has a higher capacity than ConvNeXt, and the data augmentation strategy for training the ConvNeXt models are not sufficient to harness the full potential of the CEDNet models.

\noindent\textbf{Effectiveness of the LR CED block.} Table~\ref{ablation_focal_block} shows the results of ablation experiments about the LR CED block. The model with LR CED block achieved 0.4\% box AP gains over the model without LR CED block. Since the LR CED block only incorporates a lightweight dilated \textit{depth-wise} convolution beyond the standard CED block, it introduces negligible increase in parameters (less than 1\%).

\section{Conclusion}
We present a universal network named CEDNet for dense prediction tasks. Unlike the widely-used FPN and its variants that usually employ a lightweight fusion module to fuse multi-scale features from pre-designed classification networks, CEDNet introduces several cascade stages to learn multi-scale features. By integrating multi-scale features in the early stages, CEDNet achieves more effective feature fusion. We conducted extensive experiments on several popular dense prediction tasks. The excellent performance demonstrates the effectiveness of our method.

\vspace{-2mm}
\paragraph{Acknowledgements.} We thank Chufeng Tang, Junru Tan, Kai Li, Hang Chen, Junnan Chen for valuable discussions and feedback. This work was supported by the National Natural Science Foundation of China (Nos. 62061136001, U19B2034, 61836014) and THU-Bosch JCML center.

\bibliography{iclr2024_conference}
\bibliographystyle{iclr2024_conference}

\newpage
\appendix

\section{Image Classification on ImageNet-1k}

\begin{table}[t]
  \captionof{table}{Results of image classification on the ImageNet-1K \textit{val}. All models were trained and evaluated on 224$\times$224 resolution with the same settings.}
  \label{results_imagenet}
   \centering
   \setlength{\tabcolsep}{3.0mm}{}
   \scalebox{1.0}{
    \begin{tabular}{l|cc|cc}
        \toprule
        Method & Param. & FLOPs & Top-1 Acc. \\
        \midrule
        HRNet-w32~\citep{HRNet} & 38M & 7.6G & 78.4 \\
        Swin-T~\citep{swin} & 28M & 4.4G & 81.2 \\
        MSG-T~\citep{MSG} & 25M & 3.8G & 82.4\\
        FocalNet-T~\citep{FocalNet} & 28M & 4.5G & 82.1\\
        RovCol-T~\citep{RovCol} & 30M & 4.5G & 82.2 \\

        Swin-S~\citep{swin} & 50M & 8.7G & 83.1 \\
        MSG-S~\citep{MSG} & 56M & 8.4G & 83.4\\
        FocalNet-S~\citep{FocalNet} & 50M & 8.6G & 83.4\\
        RovCol-S~\citep{RovCol} & 60M & 9.0G & 83.5 \\

        Swin-B~\citep{swin} & 88M & 15.4G & 83.4\\
        MSG-B~\citep{MSG} & 84M & 14.2G & 84.0\\
        FocalNet-B~\citep{FocalNet} & 89M & 15.4G & 83.9\\
        RovCol-B~\citep{RovCol} & 138M & 16.6G & 84.1 \\

        \midrule
        ConvNeXt-T~\citep{convnext} & 29M & 4.5G & 82.1\\
        ConvNeXt-S~\citep{convnext} & 50M & 8.7G & 83.1\\
        ConvNeXt-B~\citep{convnext} & 89M & 15.4G & 83.8\\
        \rowcolor{gray! 20} CEDNet-NeXt-T (Hourglass-style) & 34M & 5.7G & 82.6 \\
        \rowcolor{gray! 20} CEDNet-NeXt-T (UNet-style) & 34M & 5.7G & 82.9 \\
        \rowcolor{gray! 20} CEDNet-NeXt-T (FPN-style) & 34M & 5.7G & 83.1 \\
        \rowcolor{gray! 20} CEDNet-NeXt-S (FPN-style) & 55M & 9.8G & 83.9 \\
        \rowcolor{gray! 20} CEDNet-NeXt-B (FPN-style) & 95M & 16.2G & 84.3 \\
        \bottomrule
    \end{tabular}
   }
\end{table}

Table~\ref{results_imagenet} shows the results of the CEDNet models in comparison with other methods. The CEDNet models achieved better results than their counterparts, \ie the ConvNeXt models. This improvement is likely because that the pre-trained CEDNet models have slightly more parameters than their counterparts. The increased number of parameters arises from the fact that these pre-trained CEDNet models are specifically designed for downstream dense prediction tasks. In these tasks, models using CEDNet as their backbone do not need additional fusion modules, which are indispensable for their counterparts. Therefore, we allocated a few extra parameters to the CEDNet models in Table~\ref{results_imagenet} to ensure that all models in subsequent dense prediction tasks have comparable size for a fair comparison (refer to the parameters and FLOPs presented in Table 1 of the paper).
\textit{Please note that the CEDNet is specifically designed for dense prediction tasks, and surpassing state-of-the-art methods in image classification is not our goal.}

\section{Further analysis}

\begin{figure*}[t]
  \centering
  \text{\footnotesize{ConvNeXt-T \qquad \qquad \qquad CEDNet-NeXt-T \qquad \qquad \qquad ConvNeXt-T \quad \qquad \qquad CEDNet-NeXt-T}}

  \begin{minipage}[b]{0.49\textwidth}
      \centering
      \includegraphics[width=\textwidth,height=0.65\textwidth]{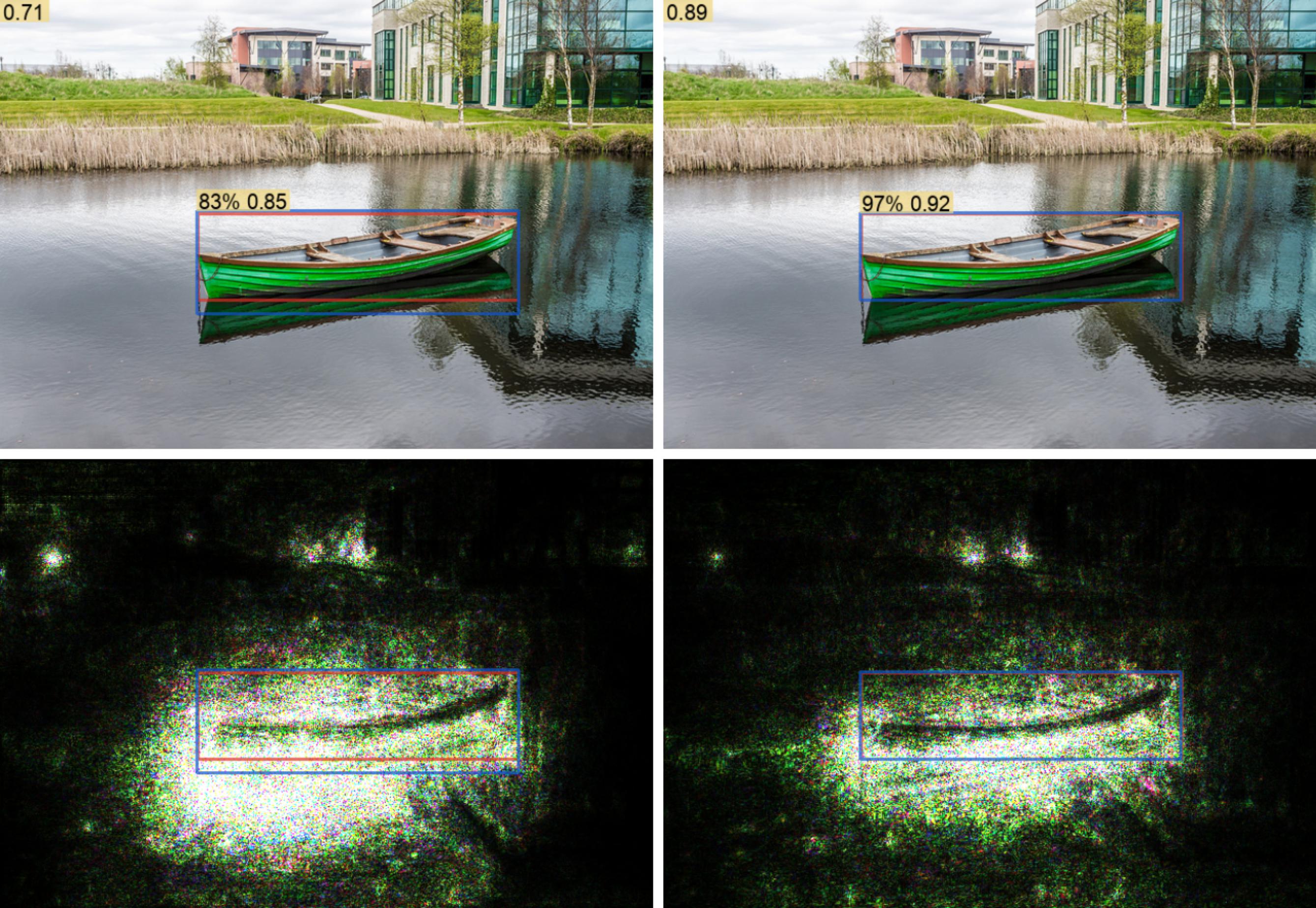}
      \text{(a)}
      \label{(a)}
  \end{minipage}
  \begin{minipage}[b]{0.49\textwidth}
    \vspace*{2mm}
      \centering
      \includegraphics[width=\textwidth,height=0.65\textwidth]{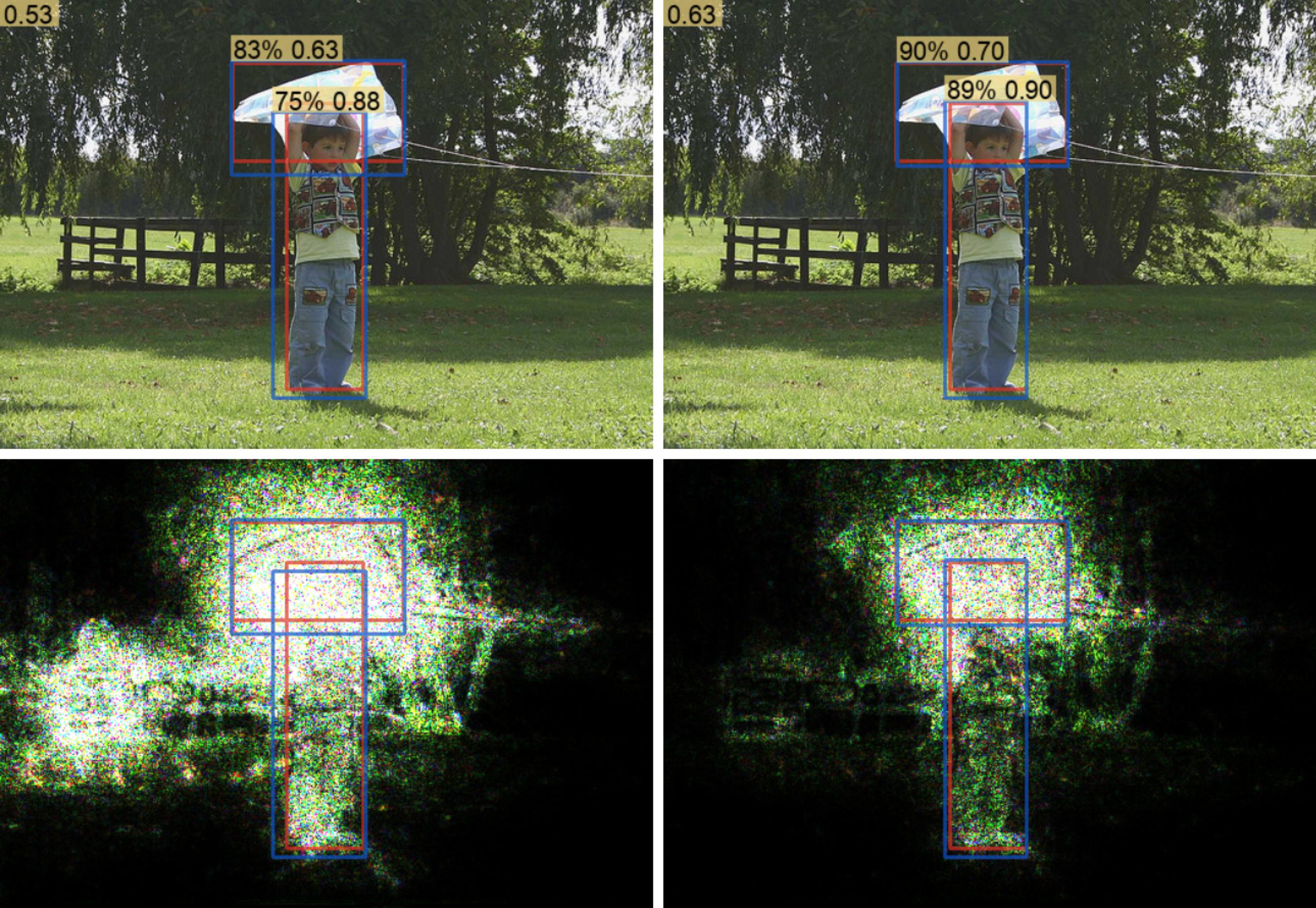}
      \text{(b)}
  \end{minipage}
  \begin{minipage}[b]{0.49\textwidth}
    \centering
    \includegraphics[width=\textwidth,height=0.65\textwidth]{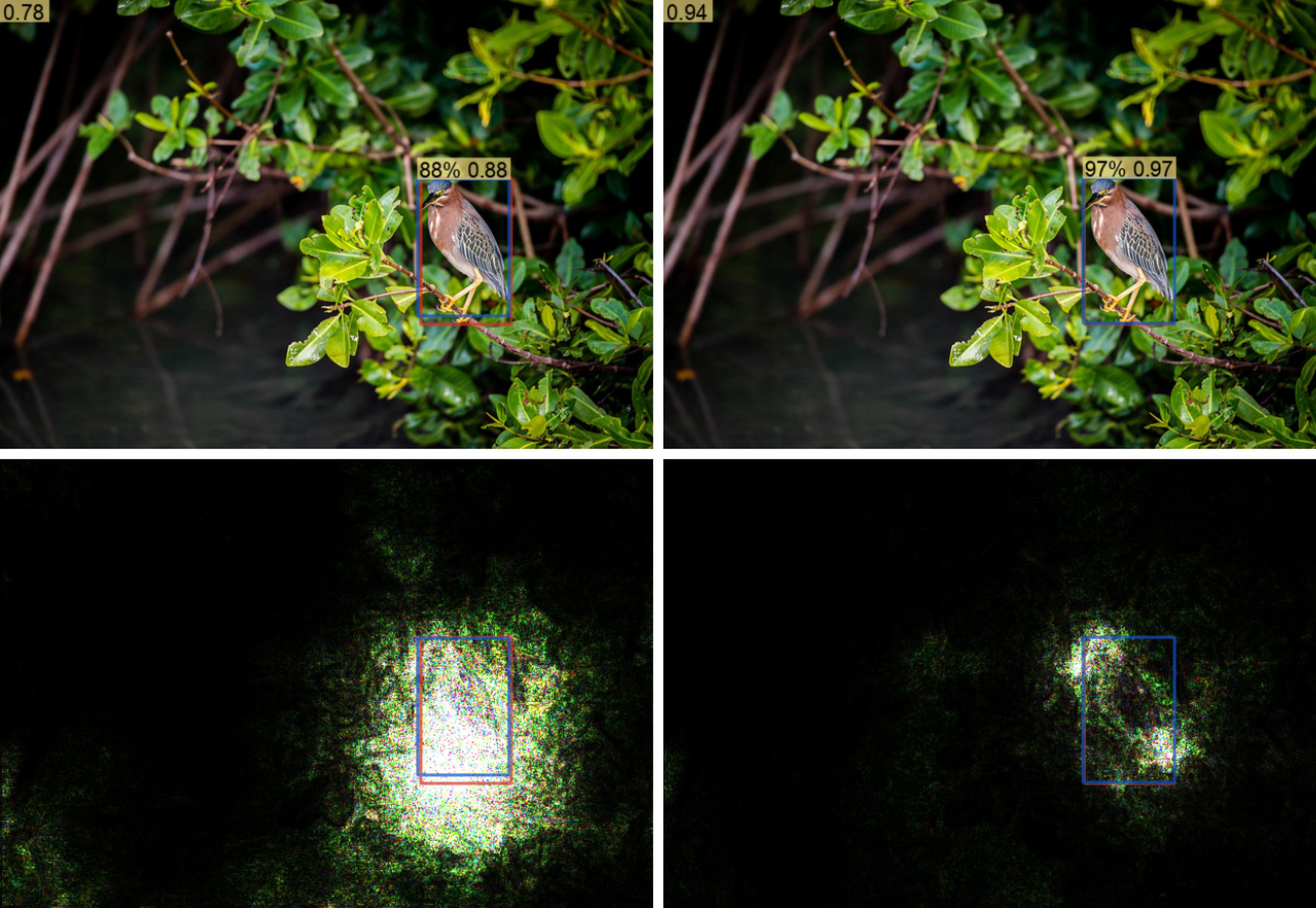}
    \text{(c)}
  \end{minipage}
  \begin{minipage}[b]{0.49\textwidth}
    \vspace*{2mm}
    \centering
    \includegraphics[width=\textwidth,height=0.65\textwidth]{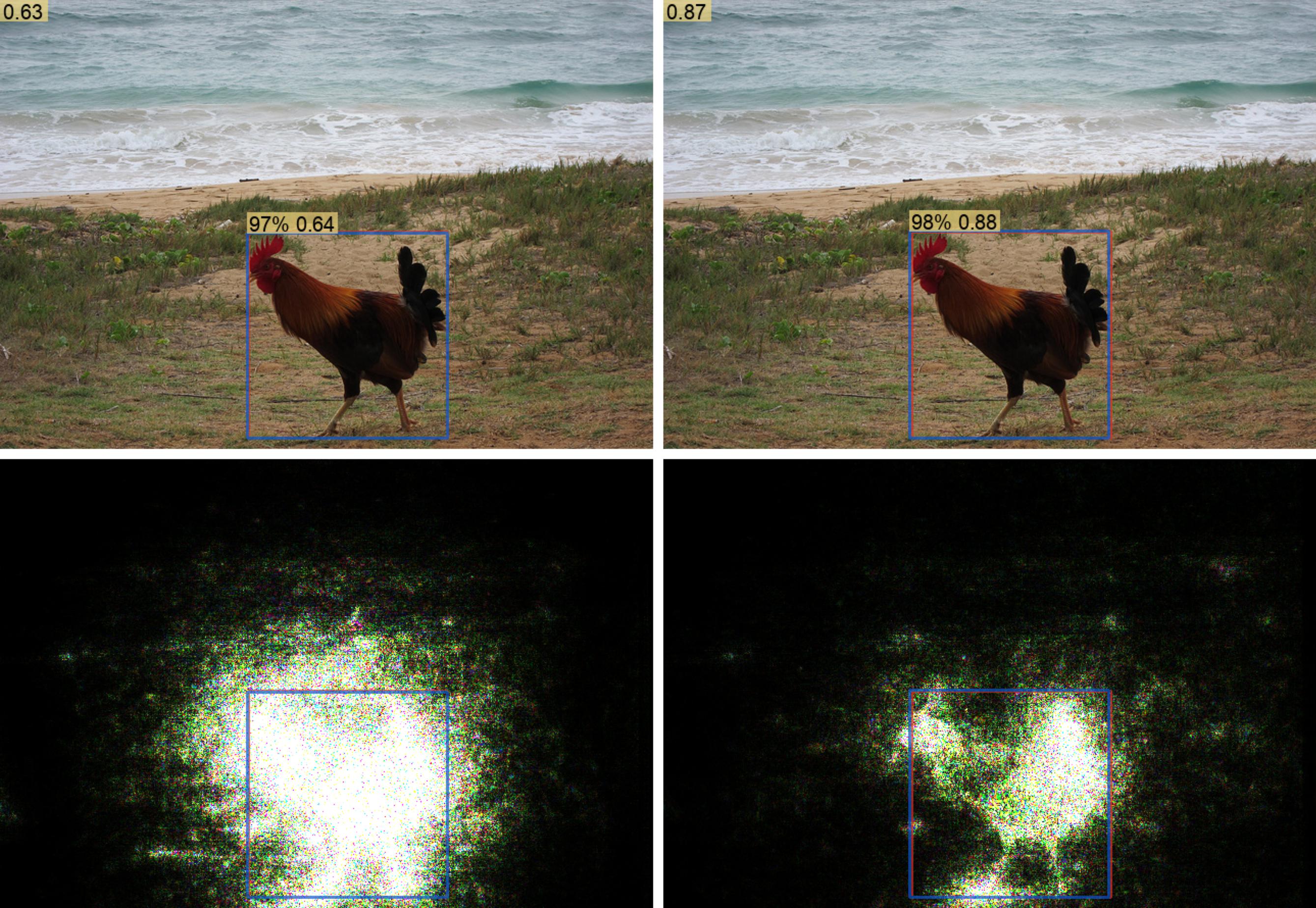}
    \text{(d)}
  \end{minipage}
  \begin{minipage}[b]{0.49\textwidth}
    \centering
    \includegraphics[width=\textwidth,height=0.65\textwidth]{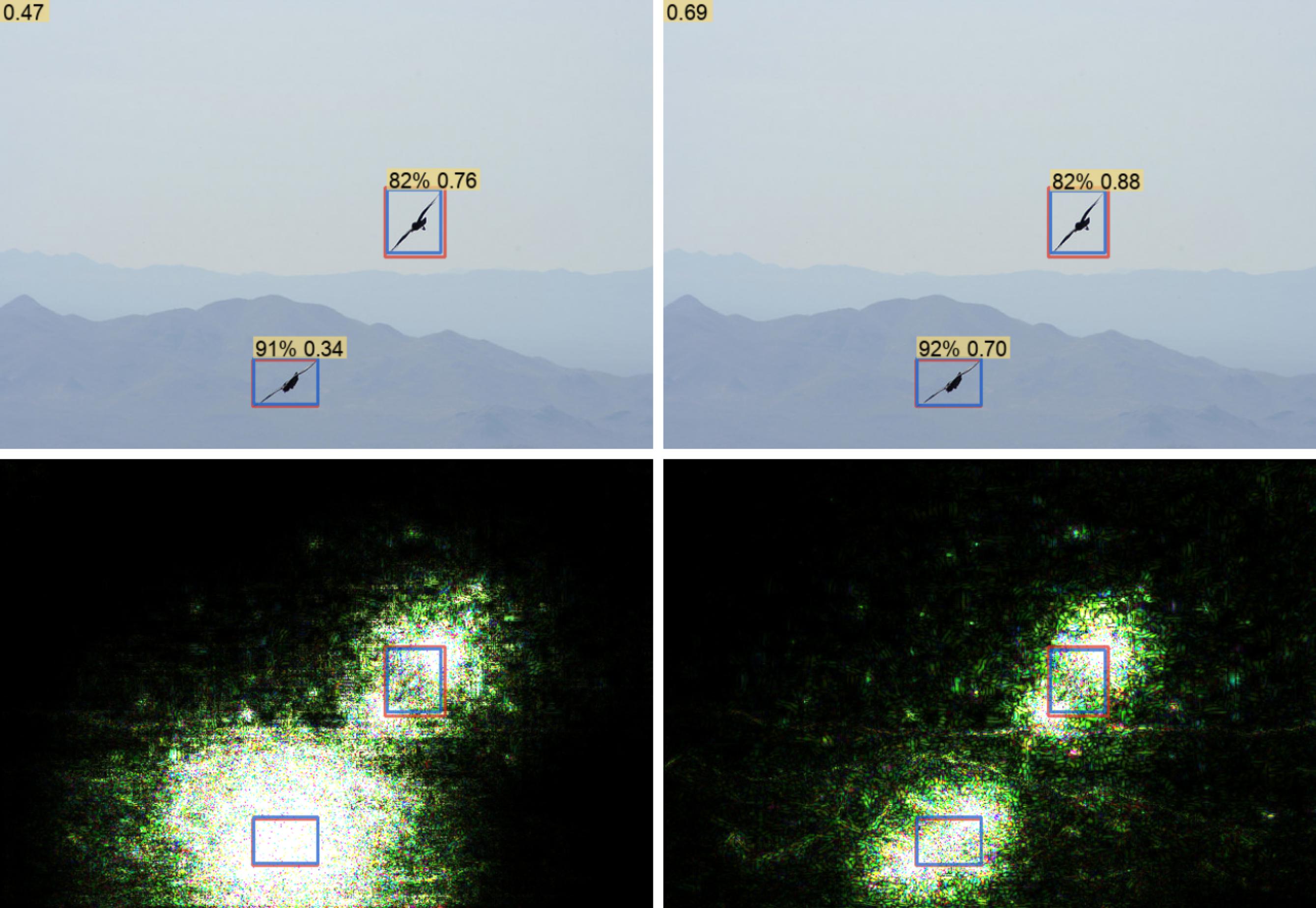}
    \text{(e)}
  \end{minipage}
  \begin{minipage}[b]{0.49\textwidth}
    \vspace*{2mm}
    \centering
    \includegraphics[width=\textwidth,height=0.65\textwidth]{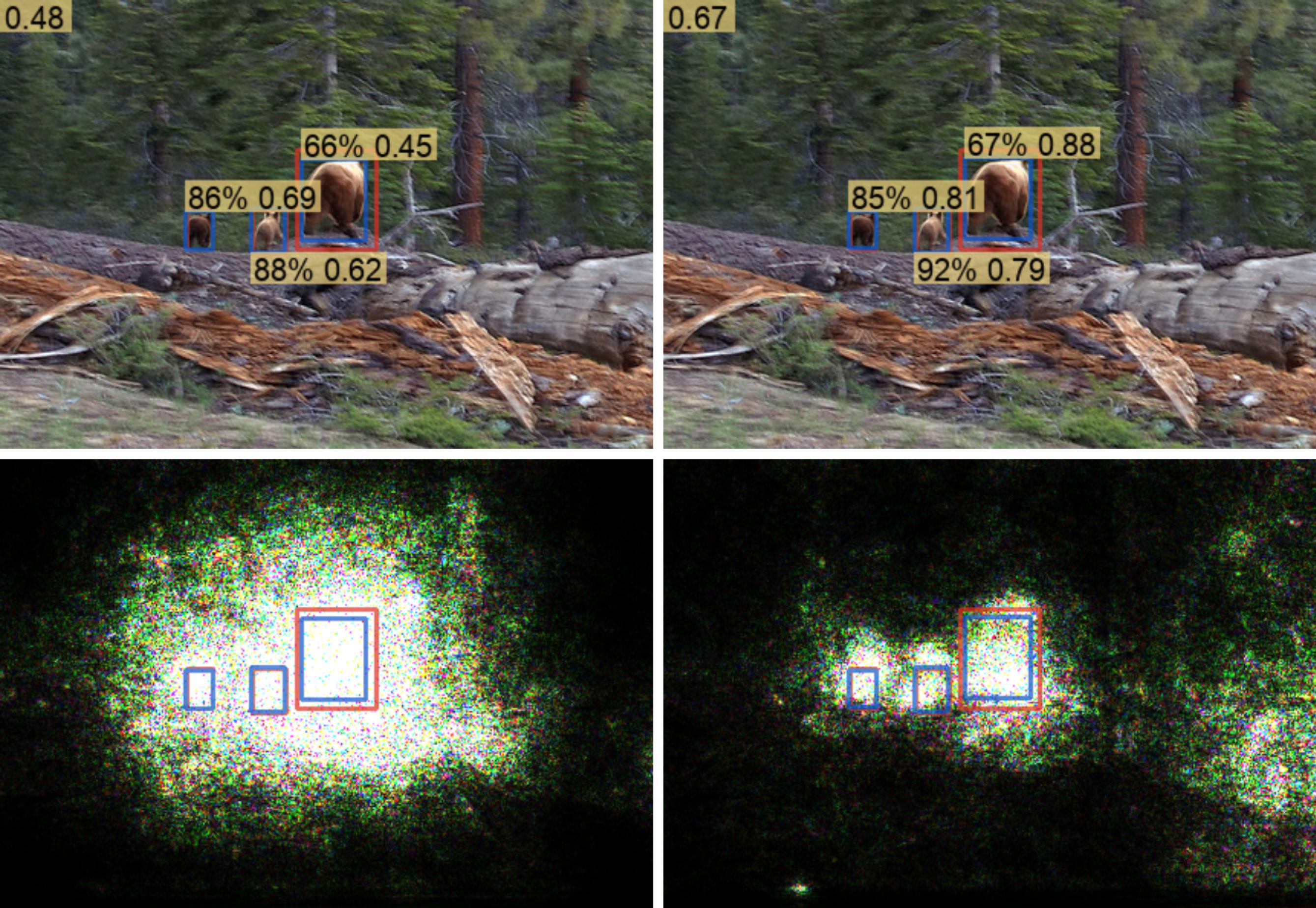}
    \text{(f)}
  \end{minipage}

  \vspace*{-2mm}
  \caption{Detection results (top row) and input gradient maps (bottom row) of RetinaNet based on ConvNeXt-T (left column) and CEDNet-NeXt-T (right column). Red boxes represent human-annotated boxes, and blue ones indicate predicted boxes. Above each predicted box, the score pair (iou, confidence) is provided. The image detection quality score is displayed in the top left corner of each image. In the gradient maps, brighter colors signify higher gradients. Please zoom in for a clearer view.}
  \label{vis_one}
\end{figure*}

We analyzed why CEDNet performed better than its baselines by comparing their input gradient distributions. Specifically, we focused on the images in the COCO \textit{val 2017} where the CEDNet models achieved the most improvements over their counterparts. We fed the selected images into the trained detectors and performed the backward process to acquire the input gradient maps of total detection loss. For each image, we define the \textit{important region} as the region where the absolute gradient is greater than $t$, and $t$ is a threshold to adjust the size of the important region.

\noindent\textbf{How to acquire the target images?} For every annotated object in an image, we first acquired the detection result with maximum IoU. If there is no matched result for a specific object, we added a virtually matched result for it and set both the IoU and confidence score of the matched result to zero. We then calculated the \textit{detection quality score} of each annotated object by multiplying the IoU and the confidence score of the matched result. The detection quality score of an image was the average quality score of all objects in the image. Finally, we obtained the top-1000 images where the CEDNet models achieved the most improvements according to the image detection quality scores.

\begin{figure*}[t]
  \centering
  \includegraphics[width=0.5\textwidth]{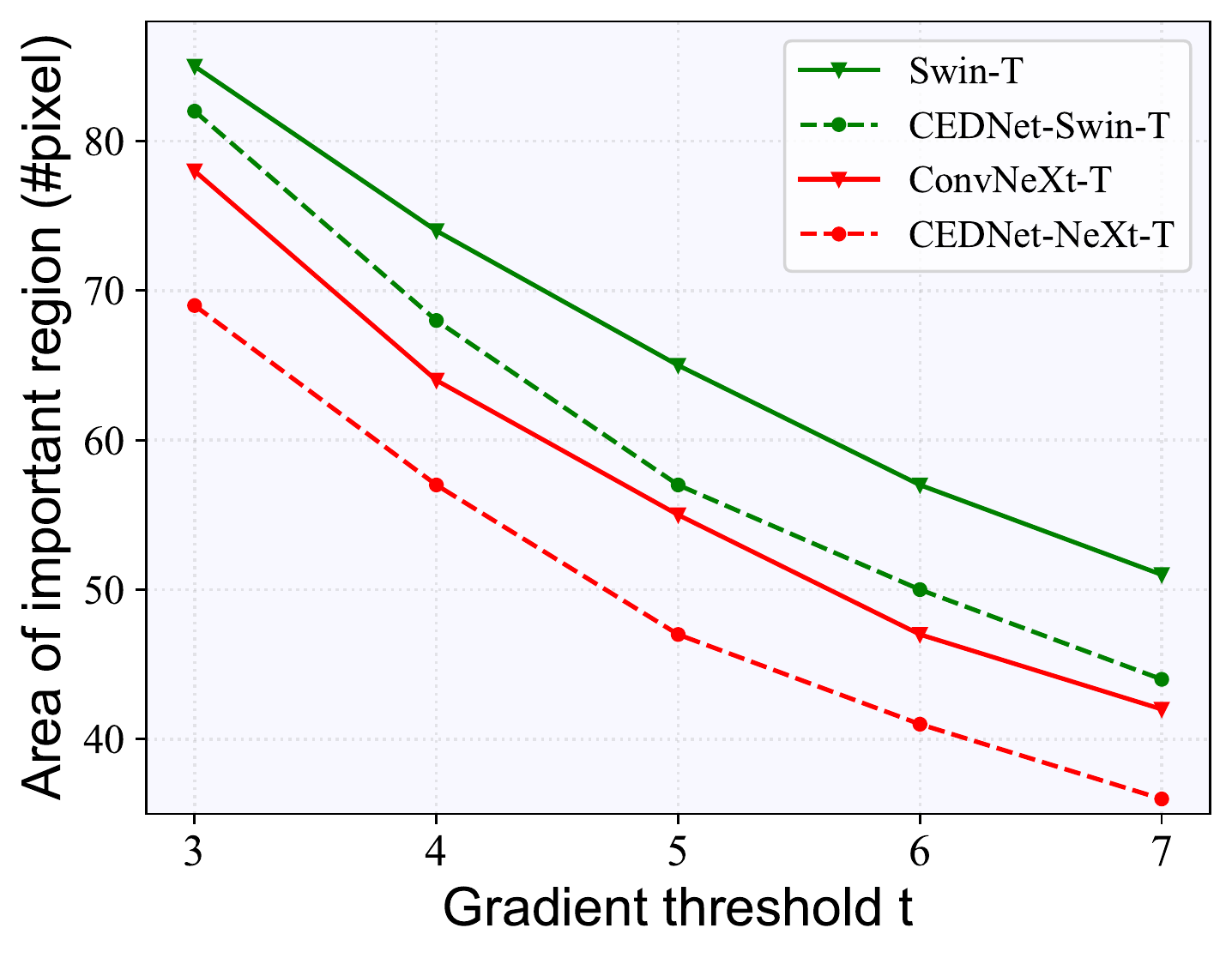}
  \vspace*{-2mm}
  \captionof{figure}{Comparison on the average area of \textit{important regions} generated from RetinaNet models based on various backbones with different gradient thresholds. We normalized the values of both axes for better visualization.}
  \label{analysis}
\end{figure*}

\noindent\textbf{Results.} We compared CEDNet with ConvNeXt and Swin Transformer based on RetinaNet. Figure~\ref{vis_one} shows that the RetinaNet with \mbox{CEDNet} concentrates more on objects with more discriminative boundaries and predicts more precise bounding boxes with higher confidences. In some cases (panels (d)-(f)), the IoUs of the boxes predicted by the CEDNet-based RetinaNet are close to that predicted by the ConvNeXt-based RetinaNet, but the confidences of these boxes predicted by the CEDNet-based RetinaNet are higher than that predicted by the ConvNeXt-based RetinaNet. The higher confidences of true positive predictions are more likely to lead to a higher mAP because the confidences of predicted boxes decide the ranking when calculating the precision-recall curve. In addition, we calculated the average area of important regions with different gradient thresholds on those selected images. Figure~\ref{analysis} shows that the average area of important regions generated from the CEDNet-based detectors is smaller than that generated from their counterparts, indicating that the CEDNet models concentrate on smaller regions.

\end{document}